%% file: main.tex
\documentclass[11pt,a4paper]{article}

\usepackage{fontspec}
\usepackage{unicode-math}
\usepackage[final]{microtype}

\usepackage{polyglossia}
\setdefaultlanguage{english}
\usepackage{csquotes}

\usepackage[margin=2.5cm]{geometry}
\usepackage{setspace}
\usepackage{parskip}
\usepackage{booktabs}
\usepackage{tabularx}
\usepackage{multirow}
\usepackage{graphicx}
\usepackage{float}
\usepackage[section]{placeins}  
\usepackage{subcaption}

\usepackage[style=apa,backend=biber,sorting=nyt]{biblatex}
\DeclareLanguageMapping{english}{english-apa}
\DeclareCaseLangs{}{}
\AtEveryBibitem{\clearfield{doi}\clearfield{url}\clearfield{isbn}\clearfield{eprint}}
\usepackage[unicode=true, breaklinks=true, colorlinks=true,
  pdfborder={0 0 0}, allcolors=linkcol]{hyperref}
\hypersetup{
  pdftitle={Mapping Political-Elite Networks in Europe with a Multilingual Joint Entity-Relation Extraction Pipeline},
  pdfauthor={Kirill Solovev and Jana Lasser}
}
\usepackage[capitalise,nameinlink,noabbrev]{cleveref}
\crefname{section}{Section}{Sections}
\crefname{subsection}{Section}{Sections}
\crefname{subsubsection}{Section}{Sections}
\crefname{figure}{Figure}{Figures}
\crefname{table}{Table}{Tables}

\usepackage{siunitx}
\usepackage{xspace}
\newcommand\eg{e.\,g.\xspace}

\usepackage{enumitem}
\setlist{nosep}
\usepackage{xcolor}
\usepackage{caption}
\DeclareCaptionFont{paper}{\fontsize{9}{11}\selectfont}
\definecolor{linkcol}{HTML}{00735C}

\title{%
  Mapping Political-Elite Networks in Europe with a Multilingual Joint Entity-Relation Extraction Pipeline%
}

\author{
    Kirill Solovev$^{1}$\footnote{Corresponding Author: \texttt{kirill.solovev@uni-graz.at}}\\
    \footnotesize{$^{1}$IDea\_Lab, University of Graz}
    \and
    Jana Lasser$^{1}$\\
    \footnotesize{$^{1}$IDea\_Lab, University of Graz}
}

\date{}

\begin{document}
\maketitle
\setcounter{footnote}{0}

\begin{abstract}
	\noindent
	A central question of comparative politics is whether political elites organise into rent-seeking coalitions that capture public resources or into civic networks that sustain governance.
	Observing the informal, economic, and adversarial ties among elites at scale has required intensive manual coding, while automated text methods have mainly measured documents or co-occurrence.
	Large language models (LLMs) can read relations from text, but cross-national use requires reproducible models, multilingual consistency, and scalable entity resolution.
	To meet these requirements, we present a modular, open-weight pipeline for multilingual joint entity-relation extraction that builds signed, temporal knowledge graphs from large news corpora.
	Span-based named-entity recognition identifies mentions, and candidate retrieval through a shared linking interface proposes identities.
	A mixture-of-experts model selects among them and extracts directed, signed relations through guided decoding against a domain ontology.
	We evaluate relation extraction against German and Polish references prepared by trained human annotators.
	In the German semantic comparison, the strongest configuration, a 397B model, reaches \qty{63.5}{\percent} LLM-adjudicated semantic F1 on 74~development articles.
	A matched-prompt comparison of that model finds nearly identical exact-label F1 with and without thinking, so thinking alone adds little in that setting.
	To assess the resulting networks, we analyse graph datasets built from \num{300000} Austrian and \num{500000} Polish articles, comparing aggregate patterns with the public record after entity deduplication and source review.
	The Austrian graph traces a party's rise, regional fracture, decline, and personnel paths into successor parties and criminal proceedings.
	The Polish graph identifies overlapping governance and economic ties around state-linked institutions and a signed conflict network between Civic Platform and Law and Justice.
	By bridging the gap between raw multilingual text and structured relational data, our framework provides a robust, replicable foundation for cross-national empirical computational social science.

	\medskip
	\noindent\textbf{Keywords:} political elite networks, large language models, joint entity-relation extraction, Wikidata, multilingual NLP, knowledge graphs, guided decoding
\end{abstract}

\clearpage

\section{Introduction}
\label{sec:intro}

Whether political elites form rent-seeking coalitions or inclusive civic networks has measurable consequences for governance \autocite{olson1982,putnam1993,acemoglu2012}.
Olson's distributional coalitions, Putnam's civic networks, and Acemoglu and Robinson's inclusive-versus-extractive framework all assign causal weight to the pattern of ties among elites in addition to their individual attributes.
Network structures shape the distribution of resources, the propagation of information, and the effects of competitive forces on the provision of public goods \autocite{knoke1990,burt1992,padgett1993}.

Large-scale research on elite networks nonetheless remains an open problem.
Structured datasets such as Archigos \autocite{goemans2009} and WhoGov \autocite{nyrup2020} cover leaders and cabinet ministers across many countries, but they only record formal positions and therefore omit informal affiliations, business ties, and adversarial relationships in the broader network.
Constructing relational data has historically required large coding teams working from biographical sources \autocite{knoke1990}, confining systematic cross-national research to a few well-resourced projects.
The \enquote{text as data} movement \autocite{hopkins2010,grimmer2013,grimmer2022} expanded what can be measured from text.
Its canonical tasks take the document as the unit of analysis, whereas network data requires relations between entity pairs within a document.

Large language models (LLMs) have begun to close this gap: they read the relational context and identify both the entities and their relationships \autocite{bro2025,lee2025,arslan2024,benoit2025}.
Cross-national use requires reproducible models, comparable extraction across languages, and entity resolution beyond study-specific name lists.
Proprietary models whose behaviour changes across application programming interface (API) versions complicate replication \autocite{chen2024chatgpt}, while evaluation in one language leaves cross-language performance untested.
Resolution against a closed roster (\eg the 155 Chilean deputies in \textcite{bro2025}) also limits which actors can be linked across studies.

To meet these cross-national requirements, we present a modular pipeline, built entirely on open-weight models, that extracts typed, entity-resolved political relationships from multilingual news.
A Simple Knowledge Organization System (SKOS) ontology of 111 entity types and 102 relationship types structures the output as cross-national, multiplex, and signed networks.
The ontology, direction constraints, and extraction guidance are separate resources that researchers can replace together for another study.
The linker accepts candidate records from Wikidata \autocite{vrandecic2014wikidata}, Orbis, and BoardEx in a shared format.
The LLM is instructed to select an identity only when the article supports the match.
Selected Wikidata identifiers (QIDs) connect mentions across languages and permit joins with structured datasets such as WhoGov, while unresolved entities remain in the output without an external identifier.
The pipeline was developed within VALPOP (\enquote{Valuing Public Goods in a Populist World}), a Horizon Europe project on how elite-network structure mediates populism, rule-of-law erosion, and public-goods provision \autocite{valpop2024}.
Its framework, shared with the sister project NET-ROL on networks and the rule of law \autocite{netrol2024,vonjacobi2025}, distinguishes Olsonian rent-seeking coalitions (O-groups) from Putnamian civic networks (P-groups).
The pipeline supplies relational data for studying both, while the case studies examine structures associated with O-groups.
This paper evaluates the measurement system needed for that comparison.
Tests of how these network structures relate to governance outcomes remain downstream applications.

We make three contributions.
(i) We present the pipeline architecture and its domain ontology, showing how modular design and ontology-constrained decoding enable adaptation across research questions, languages, and hardware (\Cref{sec:pipeline}).
(ii) We evaluate extraction against a German reference with annotations from one trained human annotator and a 100-article Polish reference based on independent work by two trained human annotators and checked against the same annotation rules (\Cref{sec:evaluation}).
These evaluations cover relation extraction alone and the complete pipeline, including comparisons of extraction instructions and verification.
(iii) We compare selected structures in the Austrian and Polish graphs with the public record (\Cref{sec:austria,sec:poland}).
Recovering these documented patterns tests whether the aggregate graphs capture structures relevant to political science.

\section{Background and Related Work}
\label{sec:background}

\textbf{Elite-network data.}
Existing structured datasets enable systematic studies of elite turnover by aggregating formal positions across countries.
For instance, Archigos \autocite{goemans2009} covers world leaders since 1875, WhoGov \autocite{nyrup2020} maps cabinet composition globally, and the World Elite Database \autocite{buhlmann2025} compiles biographical data on economic elites.
However, because these databases are restricted to formal institutional roles, they miss informal, economic, and adversarial ties in the broader political network.
Tracing these hidden connections has relied on intensive manual coding \autocite{knoke1990,keller2016}.
Automated methods fall into two categories: \emph{overlap-rule} approaches that infer ties from shared memberships and affiliations \autocite{breiger1974,fowler2006connecting,rossier2022integrated}, and \emph{extractive} approaches that infer ties from co-occurrence or relational signals in unstructured text \autocite{mahdavi2019,vanatteveldt2008,angst2025}.
The present work belongs to the extractive tradition and extracts typed, directed, and signed relationships into multiplex networks under one ontology.

\textbf{LLM-based political extraction.}
LLM-based approaches to political network extraction have become more prominent.
\textcite{bro2025} use GPT-4 to extract political networks from \num{1009} Chilean articles in a single prompt, \textcite{arslan2024} combine retrieval-augmented generation with Llama~2 for event extraction, and \textcite{lee2025} extract participant lists and source attribution with GPT.
At a similar scale but without LLMs, \textcite{tamper2022} build a knowledge graph from \num{114000} Finnish parliamentary speeches with a traditional modular NLP pipeline.
At the survey level, \textcite{heseltine2024} and \textcite{benoit2025} find that model-based coding can reach expert-level reliability.
These systems address different extraction tasks and differ in model access, language coverage, external entity linking, and output constraints (\Cref{tab:comparison}).
Our pipeline addresses these constraints through separate NER, entity-linking, and relation-extraction stages with explicit data interfaces, open-weight models, and external identity grounding.

\begin{table}[ht]
	\centering
	\caption{Representative political information-extraction systems.
		Open weights denotes the absence of proprietary API dependencies.
		Multilingual denotes applications in more than one language.
		KB-linked denotes resolution to an external knowledge base.
		Fixed ontology denotes an ontology enforced at decode time.
		Signed/temporal denotes typed positive/neutral/negative ties with temporal scope.
		Tamper et al.'s pipeline uses no generative model, so its fixed-ontology cell shows a dash.}
	\label{tab:comparison}
	\footnotesize
	\begin{tabular*}{\linewidth}{@{\extracolsep{\fill}}lccccc}
		\toprule
		\textbf{System} & \textbf{Open} & \textbf{Multi-} & \textbf{KB-} & \textbf{Fixed} & \textbf{Signed/} \\
		& \textbf{weights} & \textbf{lingual} & \textbf{linked} & \textbf{ontology} & \textbf{temporal} \\
		\midrule
		\textcite{bro2025}     & No  & No  & No  & No   & Part. \\
		\textcite{arslan2024}  & Yes & No  & No  & No   & No    \\
		\textcite{lee2025}     & No  & Yes & No  & No   & No    \\
		\textcite{tamper2022}  & Yes & No  & Yes & --   & No    \\
		\textbf{Our pipeline}        & Yes & Yes & Yes & Yes  & Yes   \\
		\bottomrule
	\end{tabular*}
\end{table}

\textbf{Entity linking.}
Multilingual entity linking is harder than the monolingual English setting \autocite{guellil2024,sevgili2022}.
Autoregressive approaches such as mGENRE \autocite{decao2022} transfer across languages but vary with training data and lexical overlap.
For political entities, Wikidata's \emph{WikiProject every politician} \autocite{wikiproject2024,everypolitician2016} maintains structured records for representatives in more than 39 legislatures, providing both a linking target and a validation source.
Our pipeline combines exact, fuzzy, and dense-vector retrieval for multilingual political text, where morphological variation (Polish has 7 grammatical cases, Hungarian 18) and cross-lingual aliases undermine the reliability of monolingual systems.
Article context guides candidate retrieval and LLM disambiguation, and anchoring resolved entities to QIDs makes their nodes language-independent and joinable with datasets such as WhoGov and WikiProject.
Wikidata's own coverage varies across languages, which leaves gaps for less prominent actors \autocite{kaffee2017}.
These constraints motivate the pipeline's separation of multilingual mention detection, candidate retrieval, and ontology-constrained relation extraction.

\section{The VALPOP Pipeline}
\label{sec:pipeline}

\subsection{Overview and Design Principles}
\label{sec:overview}

\begin{figure}[htbp]
	\centering
	\includegraphics[width=\textwidth]{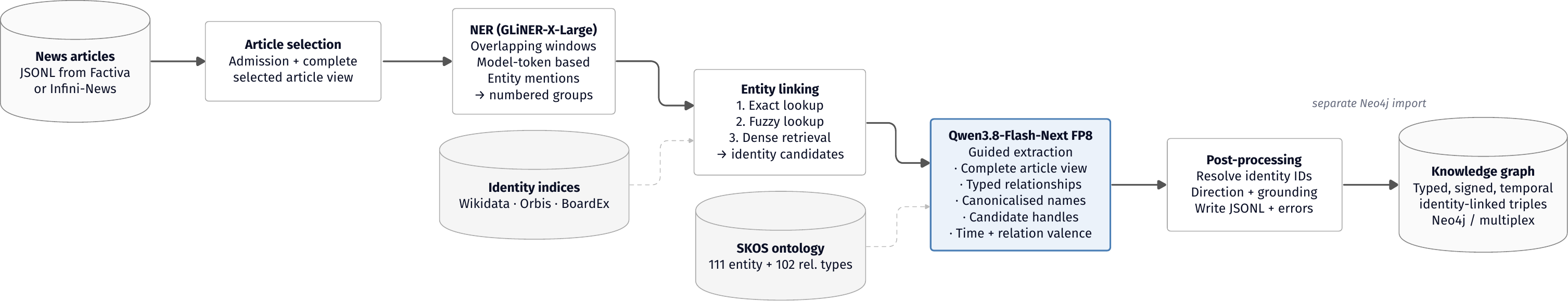}
	\caption{The pipeline's data flow.
		Article admission, windowed NER (\Cref{sec:ner}), and entity linking (\Cref{sec:el}) prepare the complete selected article view, mentions, and candidates for relation extraction (\Cref{sec:re}).
		Linking indices supply candidate identities, and the SKOS ontology (\Cref{sec:ontology}) constrains the output via guided decoding.
		Processed records are written to JSONL and imported separately into the knowledge graph.}
	\label{fig:architecture}
\end{figure}

Given an article $a$, the pipeline produces relation tuples $R(a) = \{(s_i, r_i, o_i, t_i, \sigma_i, q_i^s, q_i^o)\}$, where $s_i, o_i$ are canonicalised entity names and $r_i$ is one of the 102~relationship types in the ontology~$\mathcal{O}$.
Here, $t_i$ specifies temporal scope (event, state, or property), $\sigma_i \in \{+, 0, -\}$ gives an article-grounded cooperation--conflict sign, and $q_i^s, q_i^o$ identify the endpoints.
An endpoint identifier is a Wikidata QID, a source-prefixed registry identifier, or $\varnothing$ when identity remains unresolved.
Guided decoding fixes the type vocabulary, while extraction assigns each relation a sign and temporal scope with optional dates.
The fixed vocabulary supports comparison across runs and corpora, and the event, state, and property classes distinguish bounded events from persistent or definitional ties.

The article is the unit of processing and output.
The pipeline validates JSONL input, normalises text and metadata, and decides which articles reach the models and what text they read (\Cref{sec:admission}).
Language processing uses the declared language when recognised, otherwise a configured detection and fallback policy.
Source-country context comes only from supplied publication-region metadata.
Preprocessing applies NER and entity linking locally or on compatible GPU nodes before the separately served LLM extracts relations.
The output distinguishes skipped articles, processing failures, and successful articles with no extracted relations.

\subsection{Ontology}
\label{sec:ontology}

The ontology is VALPOP-specific, grounded in Wikidata, and implemented in SKOS \autocite{miles2009skos} with labels in 12~languages.
It comprises 111 entity types in five groups (People, Organisations, Events, Concepts, Places) and 102 relationship types in eight groups (\Cref{tab:ontology}).
The inventory combines types from the domain literature with recurrent types found in unconstrained extraction over a stratified corpus sample, after which the production vocabulary was fixed.

\begin{table}[ht]
	\centering
	\caption{VALPOP ontology structure.
		Entity and relationship types are grouped into semantic categories that define multiplex network layers.}
	\label{tab:ontology}
	\footnotesize
	\begin{tabular*}{\linewidth}{@{\extracolsep{\fill}}llrl}
		\toprule
		\textbf{Level} & \textbf{Group} & \textbf{Types} & \textbf{Examples}                           \\
		\midrule
		\multirow{5}{*}{Entity}
		& People         & 22             & politician, judge, businessperson           \\
		& Organisations  & 56             & political\_party, central\_bank, university \\
		& Events         & 16             & election, lawsuit, treaty                   \\
		& Concepts       & 12             & law, office\_or\_role, public\_contract      \\
		& Places         & 5              & city, town, sovereign\_state               \\
		\midrule
		\multirow{8}{*}{Relationship}
		& Governance     & 24             & position\_held, member\_of, represents      \\
		& Economic       & 26             & owned\_by, investor, employer               \\
		& Geographic     & 8              & headquarters\_location, country             \\
		& Personal       & 10             & spouse, educated\_at, child                 \\
		& Temporal       & 9              & inception, followed\_by                     \\
		& Descriptive    & 15             & instance\_of, award\_received               \\
		& Stance         & 4              & criticizes, statement\_disputed\_by         \\
		& Enforcement    & 6              & investigated\_by, rules\_on                 \\
		\bottomrule
	\end{tabular*}
\end{table}

The seven coarse NER labels (\Cref{sec:ner}) form a separate, coarser label set that includes locations but no explicit event label.
The LLM assigns fine-grained ontology types to named entities supported by the article and can recover entities absent from the NER inventory.
Relationship types carry directional constraints.
Nine are symmetric (\eg \texttt{spouse}), while directed relations use their defined subject--object orientation.
The ontology supplies default cooperation--conflict signs for structured records, while the LLM assigns an article-specific sign to each extracted relation (\Cref{sec:re}).

\subsection{Linking Sources and Preprocessing}
\label{sec:preprocessing}

The linking interface accepts records from Wikidata, Bureau van Dijk's Orbis, and BoardEx in a common format.
An offline pipeline processes the Wikidata JSON dump, using the P279 (subclass-of) hierarchy to identify entities within the ontology's scope.
A minimum of one sitelink retains 14.8~million entities, which are indexed for dense retrieval in Qdrant \autocite{qdrant2023} and through 37~million SQLite aliases across 36~languages.
Runtime morphological lookup supplies additional forms when direct surface-form lookup fails (\Cref{sec:el}).
Orbis and BoardEx records enter the same retrieval interface with their source identifiers, names, aliases, descriptions, and entity types.

\subsection{Article Admission}
\label{sec:admission}

News archives contain prose, notices, tables, lists, registers, headline-only stubs, and repeated link widgets.
Admission runs once per article on the CPU, before any GPU work, and decides whether the article is processed and what text the models read.
A skipped article keeps its source record and a skip reason, so downstream counts separate it from articles with zero extracted relations.

The article text joins the title, the separate \texttt{snippet} field, and the body of the source record.
Articles with fewer than 20~words are excluded.
The floor is configurable.
The models read a source-aligned view in which equal-length whitespace replaces verified embedded PDF payloads.
Masks for repeated lines and numeric blocks can optionally be enabled.
A multilingual classifier uses structural text measurements, character n-grams, and source content codes to reject non-prose in languages whose thresholds pass calibration and confirmation tests, while other languages are retained.

\subsection{Named-Entity Recognition}
\label{sec:ner}

NER uses GLiNER-X-Large \autocite{zaratiana2024}, a span-based model that accepts arbitrary labels at inference, with seven coarse labels (person, organization, location, company, bank, government, political party) at a confidence threshold of 0.5.
The backend uses language-specific Stanza tokenisation \autocite{qi2020stanza} and overlapping model-token windows, then merges predictions in article character coordinates.
Post-filters remove systematic structural and lexical false positives before grouping repeated surface forms for linking.
The linker receives mention text, coarse labels, scores, and source offsets, while the LLM assigns the 111 fine-grained ontology types.

\subsection{Entity Linking}
\label{sec:el}

Entity linking decides whether the network has one node per actor.
Without it, \enquote{Tusk}, \enquote{Donald Tusk}, and the Polish genitive \enquote{Tuska} can become separate nodes and fragment the network.
The linker retrieves candidates through exact multilingual aliases and morphological forms, fuzzy string matching \autocite{garbe2012symspell,kieras2017morfeusz}, and dense contextual search.
Candidate records share a pool across sources for model selection, with a configurable limit per mention.
Each displayed candidate has a request-local handle that identifies the exact knowledge-base record even if the model canonicalises the mention differently, and selecting 0 leaves the endpoint unresolved.
The candidate record proposes an identity, while the article supplies the evidence for selecting it and for extracting any relation.

\subsection{Relation Extraction}
\label{sec:re}

\begin{figure}[htbp]
	\centering
	\includegraphics[width=\textwidth]{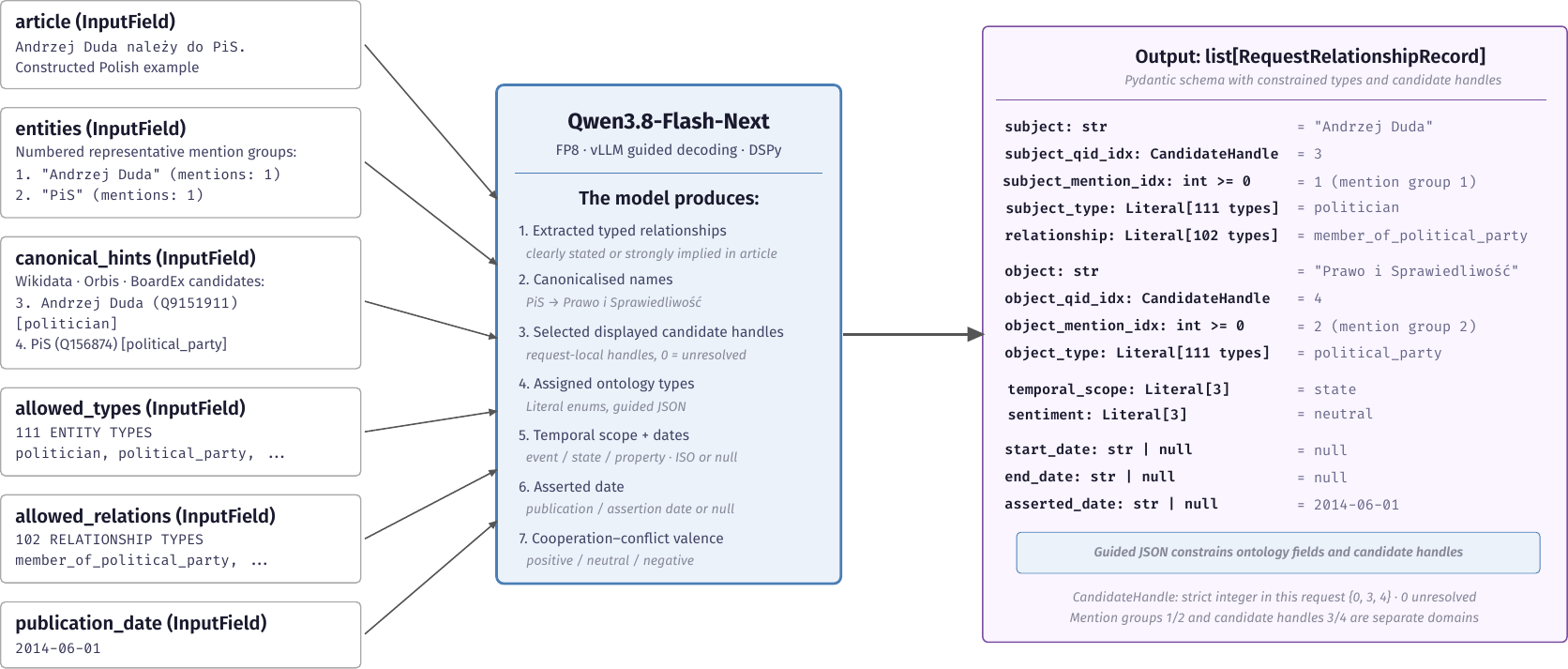}
	\caption{The relation-extraction contract, with a constructed Polish example.
		Six input fields supply the complete selected article view, numbered mention groups, candidate identity records, ontology vocabularies, and publication date.
		The relation schema has 14 fields under guided JSON decoding.
		Post-processing resolves external identifiers.
		Mention-group indices and candidate handles are separate: 0 denotes unresolved identity in the candidate fields.}
	\label{fig:dspy}
\end{figure}

Relation extraction uses Qwen3.8-Flash-Next-FP8 \autocite{qwenteam2026flashnext}, an open-weight mixture-of-experts model served via vLLM \autocite{kwon2023vllm}, with DSPy \autocite{khattab2024} and typed Pydantic signatures.
The extraction model runs in non-thinking mode and receives six input fields: the complete selected article view, numbered representative mention groups, candidate identity records, allowed entity types, allowed relationships with available guidance, and publication date.
The \texttt{allowed\_relations} resource supplies the allowed types and available guidance and can be replaced with the ontology resources for another study.
NER windows and retrieval contexts stay in preprocessing.
An article whose complete prompt exceeds the serving context limit fails with an explicit error, so every extraction reads the whole selected article text.

Each relation record has fourteen fields.
Names are canonicalised in the article's language by expanding abbreviations and resolving inflected forms.
The model assigns relationship valence from the article's framing: positive for cooperation, negative for conflict, and neutral for structural relations without either framing.
This label is stored in the \texttt{sentiment} field.
Mention-group indices attribute endpoints to the NER inventory, while candidate handles select knowledge-base identities.
Every relation must still come from the article.
JSON-schema decoding constrains entity and relationship types, post-processing validates candidate handles and assigns relationship identifiers, and direction checks correct reversible subject--object errors \autocite{geng2023grammar}.
These checks enforce output structure and ontology constraints.
The evaluation measures whether the extracted relations are supported and complete (\Cref{sec:evaluation}).

\subsection{Structured-Data Enrichment and Extensions}
\label{sec:structured}

Processed article records are written to JSONL and imported separately into a Neo4j temporal property graph, where relationship groups define multiplex layers.
External identifiers join resolved entities across articles and languages.
Unlinked entities use a key derived from their coarse ontology group and normalised, case-folded name.
The key separates people from organisations with the same name but can merge distinct same-name entities within a group.
State and property edges aggregate article evidence and relationship-valence counts, while event edges remain article-specific and provenance stays separate from substantive relations.

Structured importers add Wikidata claims, OpenSanctions, Mavise, and ICIJ Offshore Leaks records with source provenance.
They only join records through shared identifiers, using source-prefixed identifiers when no verified QID crosswalk exists.
Adding a structured source requires mapping its records, identifiers, relations, and provenance into these interfaces.

\input{evaluation_final}

The article-level evaluations measure extraction performance.
The case studies provide a complementary check: whether the reviewed corpus-scale graphs recover political structures documented independently of the pipeline.
Austria examines an organisational lifecycle and personnel trajectories, whereas Poland examines governance--economic overlap and signed partisan conflict.

\input{section_austria}

\input{section_poland}

\section{Discussion}
\label{sec:discussion}

The evaluations show that open-weight models can recover article-supported political relations, with further gains from task-specific guidance.
In the German comparison, the largest model achieves the highest semantic F1.
In a separate matched-prompt, single-seed comparison, its thinking and non-thinking configurations produce virtually identical exact-label scores.
The results therefore support selecting models through task-specific evaluation.

Instruction design offers a further route to better extraction.
In the Polish development comparison, complete relation guidance improves exact-label performance, and article-based verification raises precision while slightly reducing recall (\Cref{sec:eval-guidance-verification}).

The case studies extend the evaluation from individual relations to aggregate political structures.
After entity deduplication, targeted corrections, and source review, the Austrian graph recovers the BZÖ lifecycle and personnel trajectories.
The Polish graph recovers the signed PO--PiS cleavage and the concentration of political ties around state-linked institutions.
Agreement with these independently documented patterns supports aggregate analysis of the reviewed graphs.

Applying these analyses across countries also requires coverage of less prominent political actors in the linking knowledge base.
Requiring three Wikidata sitelinks excluded \qty{61}{\percent} of politicians listed by \emph{WikiProject every politician}.
The one-sitelink threshold retains regional and back-bench actors alongside prominent entities.

\subsection{Reproducibility}

Extending these analyses requires access to code, models, and data, a continuing challenge in computational social science \autocite{schoch2024,brodeur2024}.
The NER, embedding, and extraction models are open-weight, and the Wikidata linking index can be built from a public dump.
The ontology and reference annotations are planned for open release (\nameref{sec:Data}).
The pipeline is planned for release and is currently available for research use on request.

To support reconstruction and comparison, each run records input hashes, code and dependency identities, model revisions, ontology and admission resources, and the linking knowledge base.
Stochastic extraction, approximate retrieval, and hardware can still affect the output.

Reproducing the case-study samples requires access to the licensed Dow Jones Factiva corpus \autocite{karstens2023}.
The pipeline can also process open news corpora, including Infini-News \autocite{solovev2026infininews}, which contains over 1.3~billion Common Crawl News articles.
For reference articles also present in Common Crawl News, open-corpus identifiers can accompany the annotations without redistributing the licensed Factiva text.
The reference release will identify each sample's annotation source, evaluation role, and text-access conditions.
The ontology defines the annotation scope, so the released examples can support direct replication or adaptation to another study.

\subsection{Limitations}
\label{sec:disc-limitations}

The reported scores are development estimates.
Further evaluation should assess generalisation to unseen articles and validate model-assisted annotation and scoring through independent human review.
The admission model relies partly on model-generated labels, and its effect on downstream relation recall requires further study.

Linking coverage is higher among prominent actors than across the full set of entity nodes.
After cleanup and entity deduplication, QIDs identify \qty{19.2}{\percent} of Austrian and \qty{15.8}{\percent} of Polish entity nodes.
Missing and fragmented identities constrain network measurement, particularly for economic elites.
Business registries can supply candidates through the shared linker interface, and their effect on coverage and identity accuracy requires further study.

The timing and frequency of news reporting also shape the available relations.
Explicit start dates occur on \qty{6.9}{\percent} of Austrian and \qty{10.3}{\percent} of Polish relations, while explicit end dates occur on \qty{1.4}{\percent} and \qty{2.3}{\percent}.
Event relations carry an asserted date.
The article publication date is the only temporal anchor for the remaining undated edges.
News corpora also over-represent prominent actors and dramatic events \autocite{galtung1965,harcup2017}, so cross-national comparison requires normalisation by article volume.

\section{Ethical Considerations}
\label{sec:ethics}

The pipeline is intended for accountability research on individuals in public political and economic roles, using already-published news.
Because news also mentions private citizens, the same capacity to reconstruct interpersonal ties can support surveillance and other misuse.
We therefore review requests for access to the pipeline for research use.

The political affiliations of named individuals are special-category data under Article~9 of the General Data Protection Regulation (GDPR).
The processing rests on the scientific-research basis of Article~9(2)(j) with the Article~89 safeguards.
The study focuses on public figures acting in their public capacity and draws on licensed or publicly accessible sources.
The paper reports aggregate statistics and selected networks of named public figures.

Researchers must verify machine-extracted relations against the source text before making findings about specific people.
This review matters because an incorrect relation can make an unsupported claim about a named person.
Aggregate comparisons also require attention to coverage and language-dependent linking quality, alongside the article-volume normalisation used in our analyses.

The pipeline records ties, types, signs, and timing, which provide the measured structure for substantive interpretation.
Labelling a network \enquote{rent-seeking} is a normative interpretation of measured structure.
Keeping that interpretation distinct allows researchers to re-examine the same graph under competing theories.

\section{Conclusion}
\label{sec:conclusion}

This paper presents a modular, open-weight pipeline that extracts typed, entity-resolved political relationships from large multilingual news corpora.
Evaluation combines article-level assessment against human-derived references with case studies of independently documented political structures.
The German model comparison reaches \qty{63.5}{\percent} semantic F1 on 74~development articles.
The Polish development experiments show that relation guidance improves recall and that article-based verification raises precision while slightly reducing recall.

At the graph level, the Austrian and Polish case studies recover documented structures after entity deduplication, targeted correction, and source review.
The Austrian graph recovers the BZÖ's organisational lifecycle, the Carinthian breakaway, and personnel trajectories into successor parties.
It also maps links among politicians involved in court proceedings.
The Polish graph identifies governance--economic overlap around state-owned enterprises and maps the signed PO--PiS cleavage.
Agreement with the historical record provides an external check on these aggregate structures and supports their substantive interpretation.

Each component, including the NER model, ontology, and corpus, can be replaced through defined interfaces.
The open-weight stack supports deployment without commercial model APIs, while researchers can choose news and linking sources to match their study.
With study-specific indicators, single-country analyses can examine whether political elites form Olsonian coalitions or Putnamian civic structures.
The same framework supports cross-national comparison among countries with sufficient news and linking coverage.
Developing comparable measures of centrality, density, and network openness would connect these graphs to NET-ROL's study of societal networks and the rule of law \autocite{vonjacobi2025}.

\section*{Data Availability}
\label{sec:Data}
\addcontentsline{toc}{section}{Data Availability}

The VALPOP ontology (SKOS/Turtle) and reference annotations are planned for release under an open license as part of the VALPOP deliverables, with human and model-generated annotation sources identified separately.
The pipeline is planned for release and is currently available for research use on request.
The NER, embedding, and extraction models are publicly available: GLiNER (knowledgator/gliner-x-large), Qwen3-Embedding-0.6B, and Qwen3.8-Flash-Next-FP8.
The linking interface supports Wikidata, Orbis, and BoardEx records.
The Wikidata component is built from a public dump using documented preprocessing scripts.

\section*{Acknowledgements}
\addcontentsline{toc}{section}{Acknowledgements}

This work is funded by the European Union's Horizon Europe research and innovation programme under Grant Agreement No.~101177310 (VALPOP).
Views and opinions expressed are those of the authors and do not necessarily reflect those of the European Union or the European Research Executive Agency.

\begingroup
\setstretch{1.1}
\printbibliography[heading=bibintoc]
\endgroup

\end{document}

%% file: evaluation_final.tex
\section{Evaluation}
\label{sec:evaluation}

The evaluation asks how accurately the pipeline recovers article-supported relations and how model and instruction choices affect extraction.
We use German and Polish references created by trained human annotators.
The German comparison evaluates relation extraction from fixed preprocessing inputs, while the Polish evaluation runs the complete pipeline.

\subsection{Reference Samples and Scoring}
\label{sec:eval-reference}
\label{sec:eval-scoring}

The German collection contains 110~articles with relation annotations from one trained annotator.
The model comparison uses a fixed subset of 74~articles.

Two trained annotators independently annotated the 100~Polish articles.
Article-based model review reconciled their annotations with the current ontology.
The final reference contains \num{1094}~supported facts from \num{1738}~human annotation rows.
All 100~articles remain, including nine with no accepted relation.

Precision is the number of distinct article-supported facts divided by that number plus unsupported prediction clusters.
Supported facts count towards precision when the reference omits them.
Recall measures recovery of the reference annotations.
Semantic F1 combines precision and recall as their harmonic mean.
Semantic matching credits supported relationships with a looser or different ontology label, while exact-label scores require the correct relation type.
These scores assess relation extraction, with identity, relationship valence, and date accuracy outside their scope.

\subsection{Model Choice}
\label{sec:eval-models}

The German comparison supplies the same stored article, entity, and candidate inputs to each extraction configuration.
All configurations use an instruction that requests clause-by-clause coverage and every supported role relation, with explicit handling of negation and modality.
Each configuration is scored against the same 526~article-supported reference rows.

\begin{table}[htbp]
    \centering
    \caption{Semantic relation extraction on 74~German development articles, judged against the source text.
        Recall uses 526~common accepted reference rows.
        Counts pool two extraction seeds for the 397B model and three for each other configuration.
        Serving and model-review settings also vary across configurations.}
    \label{tab:semantic-models}
    \footnotesize
    \begin{tabular*}{\linewidth}{@{\extracolsep{\fill}}llrrr}
        \toprule
        \textbf{Model} & \textbf{Configuration} & \textbf{Precision} & \textbf{Recall} & \textbf{F1} \\
        \midrule
        Qwen3.5-397B-A17B & Thinking & 0.778 & 0.537 & 0.635 \\
        Qwen3.6-35B-A3B-FP8 & Non-thinking, distributed & 0.713 & 0.515 & 0.598 \\
        Qwen3.6-35B-A3B-FP8 & Non-thinking, single GPU & 0.682 & 0.508 & 0.582 \\
        Qwen3.8-27B & Non-thinking & 0.797 & 0.449 & 0.574 \\
        \bottomrule
    \end{tabular*}
\end{table}

The 397B thinking configuration reaches the highest semantic F1, \qty{63.5}{\percent}, compared with \qty{59.8}{\percent} for the strongest 35B configuration (\Cref{tab:semantic-models}).
The 27B configuration has the highest precision and lowest recall.

A matched single-seed comparison on 110~German articles using the same 397B model and original instruction found automatic exact-match F1 of 0.3101 with thinking and 0.3096 without it.
Thinking mode therefore produced almost no F1 change in that comparison.

\subsection{Complete-Pipeline Performance}
\label{sec:eval-polish}

The Polish evaluation runs entity recognition, candidate retrieval, and relation extraction, with Qwen3.8-Flash-Next-FP8 at the relation-extraction stage.

The complete pipeline reaches semantic precision of \qty{57.4}{\percent}, fact recall of \qty{49.3}{\percent}, and fact-level F1 of \qty{53.0}{\percent} (\qty{95}{\percent} article-bootstrap interval: \qtyrange{49.0}{56.8}{\percent}).
The interval uses \num{100000}~article-bootstrap draws.
Of its 725~supported facts, 186 (\qty{25.7}{\percent}) are absent from the reference.
Exact-label F1 is lower, at \qty{35.2}{\percent}: 175~supported prediction clusters reach a reference fact under a different relation label.

\subsection{Improving Coverage and Precision}
\label{sec:eval-prompt-development}
\label{sec:eval-guidance-verification}

Revised relation guidance and article-based verification address different extraction errors.
The experimental workflow supplies definitions and direction hints for all 102~relation types, then asks the extraction model to review its predicted claims against the article.
Verification removes claims judged unsupported by the article.

On 20~Polish development articles, revised guidance followed by verification raises observed exact-label hybrid F1 from \qty{37.1}{\percent} to \qty{50.2}{\percent} (\Cref{tab:guidance-verification}).
The hybrid metric combines precision over distinct article-supported facts and error clusters with recall over the block's 273~human-reference facts.
Guidance raises recall, while verification raises precision and slightly reduces recall.

\begin{table}[htbp]
    \centering
    \caption{Guidance and verification on 20~Polish development articles, using shared article-based model judgments and 273~reference facts.
        The values describe the observed block, with hybrid F1 combining article-supported-fact precision and human-reference-fact recall.}
    \label{tab:guidance-verification}
    \footnotesize
    \begin{tabular*}{\linewidth}{@{\extracolsep{\fill}}lrrr}
        \toprule
        \textbf{Configuration} & \textbf{Precision} & \textbf{Recall} & \textbf{Hybrid F1} \\
        \midrule
        Coverage instruction & 0.437 & 0.322 & 0.371 \\
        Revised guidance & 0.558 & 0.418 & 0.478 \\
        Revised guidance + verification & 0.721 & 0.385 & 0.502 \\
        \bottomrule
    \end{tabular*}
\end{table}

%% file: section_austria.tex
\section{Party Lifecycle in Austrian News}
\label{sec:austria}

The Austrian study uses the resulting graph for two tasks: temporal coverage and entity tracking from mentions, and political-network measurement from typed relations.
Before analysis, we merged name variants into canonical party and person nodes, verified the party nodes against Wikidata, defined the analytic cohort, and reviewed source articles to correct identity and relation-direction errors.
For the BZÖ cohort, the median rate of investigation, adjudication, and disputed-statement edge support per mentioning article is 6.9$\times$ the matched-control rate (\Cref{sec:austria-scandal}).
The temporal analysis begins with the party's 4~April 2005 secession from the Freedom Party of Austria (Freiheitliche Partei Österreichs, FPÖ), led by Jörg Haider, and his fatal car crash on 11~October 2008.
It then covers the December~2009 Carinthian breakaway as the Freiheitliche Partei Kärntens (FPK) after the nationalisation of the Hypo Group Alpe Adria bank.
The BZÖ lost all federal mandates in September~2013.
We compare these structures with the documented chronology of the Austrian far-right tradition \autocite{heinisch2008, luther2011, pelinka2017, mueller1992}, Wikidata's party-membership records, and court convictions.
These comparisons assess whether the corrected graph contains known structures.

\subsection{Corpus and Graph}
\label{sec:austria-graph}

The year-by-source proportional sample contains \num{300000}~Austrian-press Factiva articles from April 2005 through 2017, of which \num{266754} pass admission.
After post-import cleanup, the graph contains \num{485523}~entity nodes and \num{967582}~relations across 101~relation types.
Because only \qty{19.2}{\percent} of entity nodes carry a Wikidata QID, party-name surface variants remain separate nodes.
Before macro-level aggregation, we consolidate these variants into Wikidata-verified canonical party nodes (nine parties, 38~surface variants) and keep regional sub-parties, parliamentary clubs, and foreign homonyms as separate nodes.
For personnel analysis, we define a 29-person cohort of BZÖ affiliates, each keyed on extracted node names.
We merge 68~name-keyed fragments (titled and abbreviated forms of a member's full name) into the canonical person nodes, while bare surnames and mentions of relatives or staff are held out.
Manual review confirmed that each cohort node represented a single person.

\subsection{Lifecycle and Founder Death}
\label{sec:austria-lifecycle}

\begin{figure}[htbp]
	\centering
	\includegraphics[width=\textwidth]{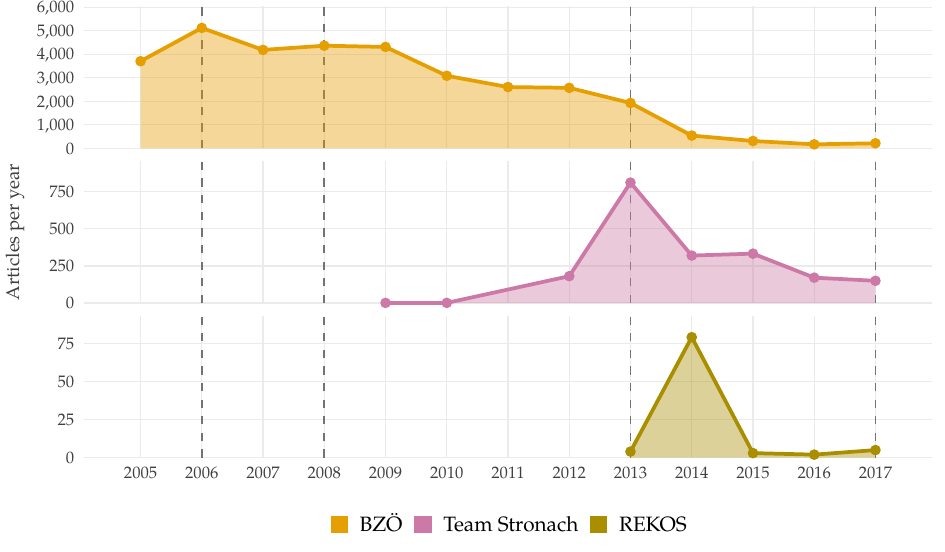}
	\caption{Annual press volume of three Austrian populist projects: BZÖ, Team Stronach, and Die Reformkonservativen (REKOS). Vertical dashed lines mark Nationalrat elections.
		The series captures major BZÖ milestones from the 2005 secession to the 2013 federal electoral defeat, with Team Stronach emerging as a successor project.}
	\label{fig:austria-lifecycle}
\end{figure}

The annual mention series captures the major lifecycle milestones in press coverage (\Cref{fig:austria-lifecycle}).
Extracted coverage emerges abruptly with the 2005 secession (\num{3703}~mentions from April onwards) and peaks in the 2006 election year (\num{5118}).
It plateaus through the 2008 election and the 2009 Hypo nationalisation and FPK split (\num{4184}--\num{4362}/year), then declines monotonically through 2013 (\num{3084}~$\rightarrow$~\num{1929}).
Following the September~2013 federal electoral defeat, coverage collapses and falls 8.9$\times$ between 2013 and 2017.
This decline also provides a comparison with Team Stronach as a successor project: near-zero presence until its 2012 founding, a spike at the 2013 Nationalrat election (812~mentions), and fading as the BZÖ collapses.

\begin{figure}[htbp]
	\centering
	\includegraphics[width=\textwidth]{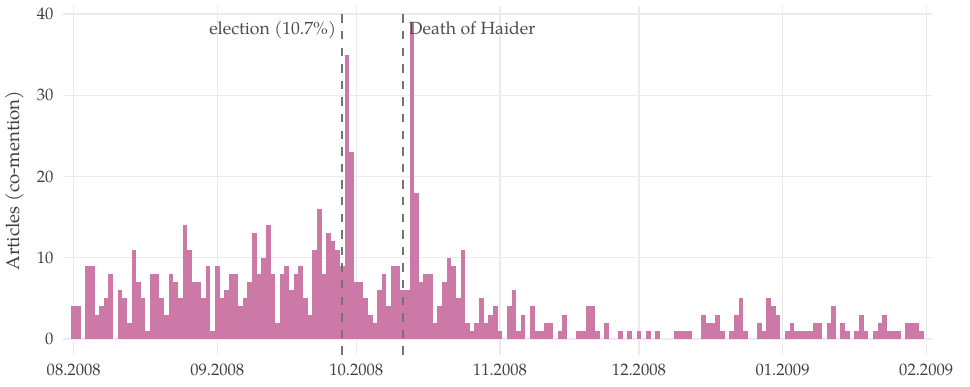}
	\caption{Daily Haider--BZÖ co-mentions around the founder's death.
		Haider died at roughly 01:00 on Saturday 11.10.2008.
		The first business-day cycle peaks at $n=39$ co-mentions on Monday 13.10.2008.}
	\label{fig:austria-death}
\end{figure}

Daily Haider--BZÖ co-mentions show a first-business-day peak after Haider's death (\Cref{fig:austria-death}).
The series records low weekend press volume ($n=6$ on both Saturday and Sunday) followed by a sharp increase, peaking at $n=39$ co-mentions on Monday, 13~October.
Haider's car crash occurred at about 01:00 on Saturday, 11~October 2008.
The Monday peak is the expected signature of a weekend event in a daily press cycle.

\subsection{The Hypo--FPK Fracture}
\label{sec:austria-hypo}

\begin{figure}[htbp]
	\centering
	\includegraphics[width=\textwidth]{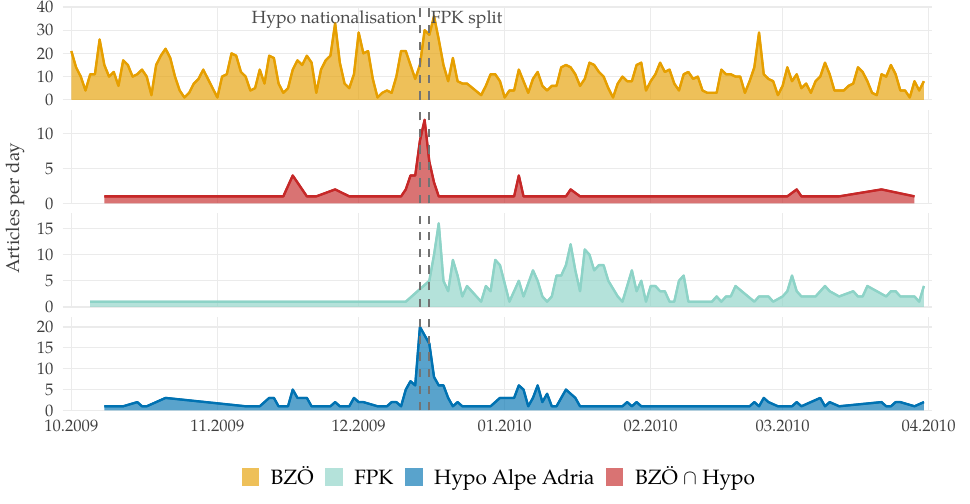}
	\caption{The Hypo$\rightarrow$FPK fracture, 12.2009.
		Daily mentions of BZÖ, FPK, the Hypo bank, and the BZÖ$\,\cap\,$Hypo intersection.
		Dashed lines mark the emergency nationalisation (14.12.2009) and the FPK breakaway announcement (16.12.2009), two days apart.
		The BZÖ federal reaction peaks at $n=36$ on 17.12.2009, matching the documented chronology.}
	\label{fig:austria-hypo}
\end{figure}

The daily series also follows the documented sequence of the regional fracture (\Cref{fig:austria-hypo}).
Graph intersections between BZÖ, the breakaway FPK, and the Hypo Group Alpe Adria bank spike sequentially in mid-December 2009.
Hypo coverage surges on December 14--16 ($n=20/18/16$, against 5--7 articles per day in the preceding week).
FPK volume steps up from ${\sim}$0--1 to 5, 10, and 16 articles per day on December 16--18.
The BZÖ federal reaction peaks at $n=36$ on December 17.
The dates align with the historical record: the Republic emergency-nationalised the bank on 14~December 2009, and the Carinthian branch announced its FPK breakaway two days later.
The press series rises around the December announcement and the January convention.

\subsection{Personnel Dispersal}
\label{sec:austria-diaspora}

\begin{figure}[htbp]
	\centering
	\includegraphics[width=\textwidth]{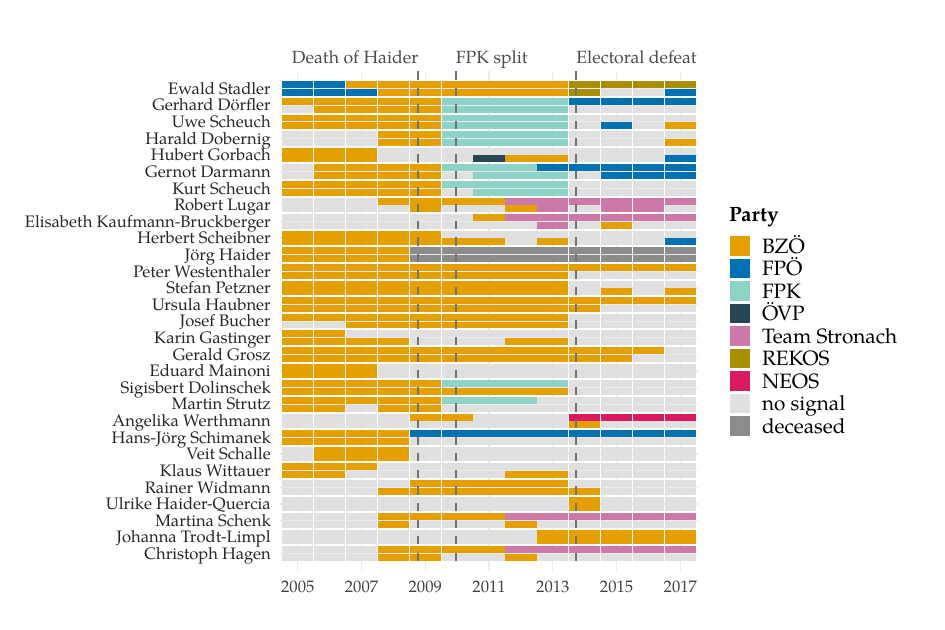}
	\caption{Dispersal of the 29-person BZÖ cohort.
		Each row is one person, and each yearly cell is split.
		The top half is curated party membership (Wikipedia and Wikidata P102), and the bottom half is the pipeline's dominant co-mention party in the Austrian press.
		A two-tone cell marks divergence between the co-mention party and formal membership.
		Vertical dashed lines mark Haider's death (2008), the FPK split (2009), and the 2013 electoral defeat.}
	\label{fig:austria-diaspora}
\end{figure}

The personnel co-mention analysis follows the cohort through the FPK split and its dispersal after the 2013 defeat (\Cref{fig:austria-diaspora}).
Through 2005--2009, the whole 29-person cohort is associated mainly with the BZÖ.
The highest observed annual switch rate occurs in 2010: 3 of 17 at-risk members switch their dominant party association (\qty{17.6}{\percent}, compared to zero in every preceding year), and two further Carinthians follow in 2011.
Departures after the 2013 defeat spread over 2013--2017 (one switch per year in 2013 and 2014), and a loyalist core keeps the BZÖ as its dominant co-mention party.

The co-mention series thus tracks personnel from a defunct party into its successors.
The 2010--2011 switches identify the Carinthian core (Gerhard Dörfler, Harald Dobernig, and both Scheuch brothers, joined by Gernot Darmann) moving to the FPK.
The series then tracks their path back into the FPÖ as the FPK is reabsorbed in June~2013, with Darmann's dominant association reverting to the FPÖ from 2015.
The series shows Robert Lugar and Elisabeth Kaufmann-Bruckberger moving to Team Stronach in 2013 and Ewald Stadler moving to REKOS in 2014.
The loyalist core, including the long-serving federal chair Josef Bucher, Peter Westenthaler, Gerald Grosz, and Stefan Petzner, keeps the BZÖ as its dominant co-mention party.
These co-mention trajectories correspond to the documented record.

\begin{figure}[htbp]
	\centering
	\includegraphics[width=0.75\textwidth,height=65mm,keepaspectratio]{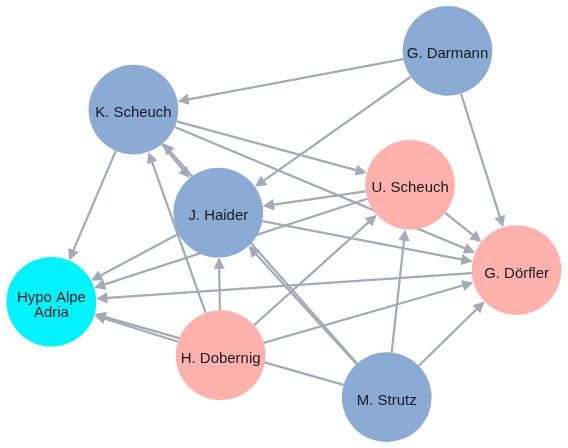}
	\caption{The Carinthia/Hypo bloc identified by Louvain on the BZÖ ego-network.
		Coral nodes identify Gerhard Dörfler, Uwe Scheuch, and Harald Dobernig, who were convicted in the cited Austrian proceedings.
		Blue nodes identify the remaining core members.
		The cyan bank display node groups the two linked bank entities contributing relations to this cohort.
		Each grey tie combines the extracted relations between two nodes in both directions, so its arrowhead does not show who acted on whom.}
	\label{fig:austria-bloc}
\end{figure}

\begin{figure}[htbp]
	\centering
	\includegraphics[width=\textwidth]{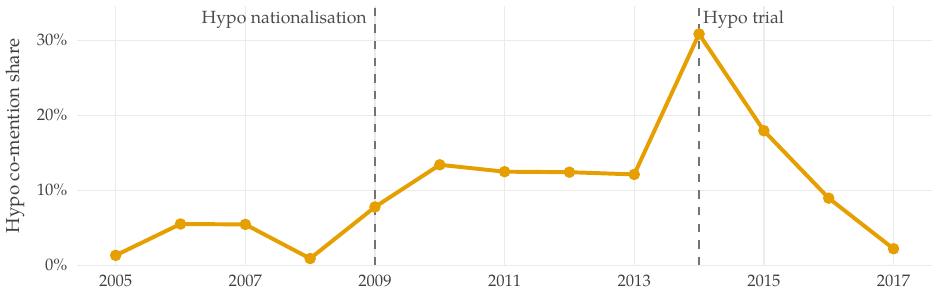}
	\caption{Haider's share of co-mentions tied to the Hypo bank over time, rising from \qty{1.3}{\percent} in 2005 to \qty{30.8}{\percent} during the 2014 trials before receding to \qty{2.2}{\percent} in 2017.}
	\label{fig:austria-hypoconc}
\end{figure}

The network also clusters several actors later involved in the cohort's criminal proceedings.
A Graph Data Science (GDS) Louvain \autocite{blondel2008louvain} partition of the BZÖ ego-network (\num{5705}~nodes, eight communities, modularity~0.27) places the Carinthian core (Dörfler, the Scheuch brothers, Dobernig, Strutz), Haider, Petzner, the FPK, and the Hypo bank in one community (\Cref{fig:austria-bloc}).
At the default resolution, a seeded Leiden \autocite{traag2019leiden} partition also groups these actors and the bank, while Gernot Darmann remains outside the group.
After removing the nine surname-fragment nodes, the Carinthian personnel core remains together, while the Hypo bank and Darmann fall into separate communities.
At lower resolution the bank joins a neighbouring institutional community, and at higher resolution the personnel core separates from Haider.
The Carinthian personnel core is therefore stable across these specifications.
The community also contains actors convicted in separate proceedings.
The €20~billion in state guarantees run up during the Haider era culminated in a €4 nationalisation.
Three community members were later convicted in the cited proceedings: governor Gerhard Dörfler (breach of trust over a state-funded BZÖ election brochure \autocite{doerfler2018wahlbroschuere}), deputy governor Uwe Scheuch (corruption \autocite{scheuch2013bribery}), and finance councillor Harald Dobernig (breach of trust \autocite{dobernig2016untreue}).
The bank's inflated-fee \enquote{Birnbacher} sale produced a further breach-of-trust conviction, of the Carinthian ÖVP leader Josef Martinz \autocite{birnbacher2014ogh}, who sits outside the BZÖ network.
Haider's share of co-mentions with the Hypo bank follows the legal aftermath.
It rises from \qty{1.3}{\percent} in 2005 to \qty{7.8}{\percent} in the nationalisation year, \qty{13.4}{\percent} in 2010, and \qty{30.8}{\percent} during the 2014 trials.
The share recedes to \qty{2.2}{\percent} in 2017 (\Cref{fig:austria-hypoconc}).

\subsection{Extracted Investigation, Adjudication, and Dispute Relations}
\label{sec:austria-scandal}

\begin{figure}[htbp]
	\centering
	\includegraphics[width=\textwidth]{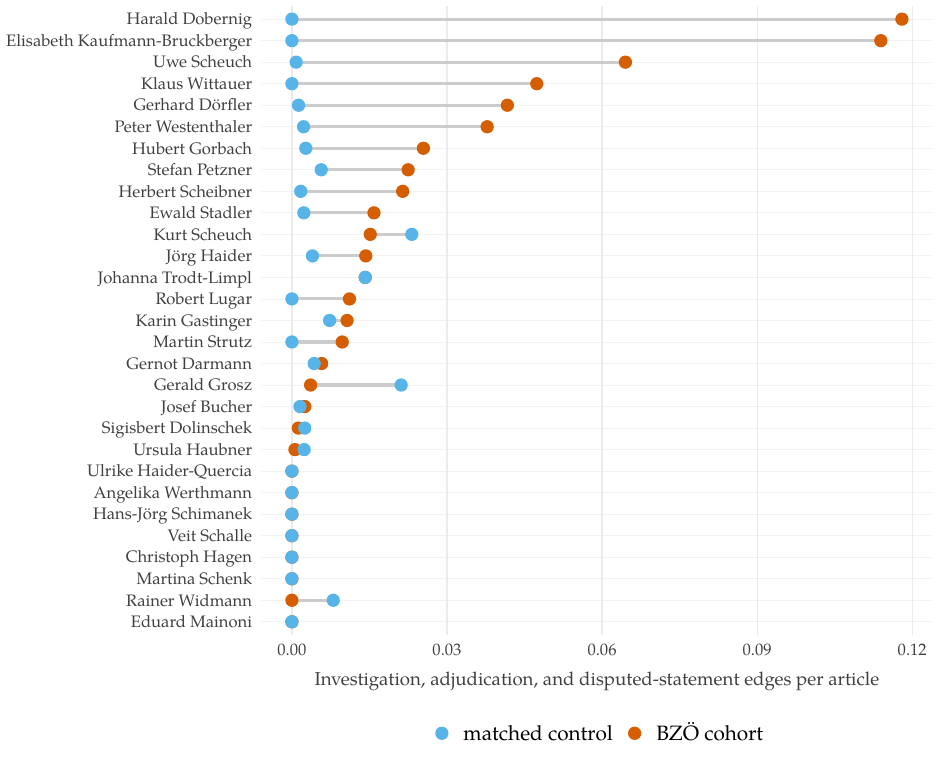}
	\caption{Selected-relation density of the BZÖ cohort against a 1:1 press-matched, Wikidata-verified control.
		On investigation, adjudication, and disputed-statement edges per article, the cohort runs at 6.9$\times$ the control (paired one-sided Wilcoxon $p=0.002$).}
	\label{fig:austria-scandal}
\end{figure}

The extracted edges allow a comparison of selected relation density between the BZÖ cohort and a political baseline (\Cref{fig:austria-scandal}).
We construct a 1:1 Wikidata-verified control group matched on Austrian press presence, consisting of non-BZÖ office-holders.
The control includes high-presence FPÖ figures alongside low-profile office-holders.
We define the three-type subset as any incident \path{investigated_by}, \path{rules_on}, or \path{statement_disputed_by} relation in either direction per Austrian article mentioning that person, excluding the dominant \texttt{criticizes} relation.
On this subset, the BZÖ cohort runs at 6.9$\times$ the median density of the control, with the cohort member ahead in 16 of the 29 matched pairs (paired one-sided Wilcoxon signed-rank $p=0.002$, and one-sided Mann--Whitney $p=0.005$).
The share of articles carrying at least one edge of these three types gives the same picture (6.3$\times$, one-sided Wilcoxon $p=0.002$).
Expanding to a four-type measure that includes \texttt{criticizes} shrinks the gap to 1.6$\times$ (paired one-sided Wilcoxon $p=0.007$, and one-sided Mann--Whitney $p=0.04$).
The difference between the groups is largest for investigation, adjudication, and disputed-statement relations.
Adding criticism reduces this difference.

\subsection{Validation and Limits}
\label{sec:austria-validation}

Against Wikidata's date-qualified P102 party-membership statements, the year-by-year trajectory makes zero false assignments on the 16 cells for which it assigns a party (precision \num{1.000}) and abstains on the remaining 10 cells (recall 0.615).
Only 3~of the 29~cohort members have date-qualified P102 statements, so this check is illustrative and the remaining dated assignments require manual verification.
Topologically, Jörg Haider is the most central actor around the BZÖ on all three centrality measures.
He leads distinct-neighbour degree (\num{2772}, 2.2$\times$ the next person) and weighted PageRank on the full content graph (2.3\textperthousand{} of the mass, 2.3$\times$ Heinz-Christian Strache and 2.9$\times$ the next BZÖ figure, Peter Westenthaler).
Haider also leads exact betweenness on the BZÖ ego-network (4.0$\times$ the runner-up, Strache), a result that holds after removing all mislinked surname nodes.

The dominant-party signal is a press co-mention proxy that lags formal membership by approximately one year.
Office-extraction precision also remains low (\qty{18.5}{\percent} per person-office pair against Wikidata P39, \qty{24.4}{\percent} per edge), driven by generic office labels and by models mistaking candidacy for held office.
Entity fragmentation persists at the person and organisational levels: nine family-name items in the BZÖ ego-network carry real co-mention degree, although they account for only \qty{0.03}{\percent} of PageRank mass.
These items are flagged and excluded from substantive analysis, and bare-surname mentions (for example \enquote{Grosz} or \enquote{Dörfler}) stay unattributed and outside the cohort.
Before deduplication, the Hypo bank's mentions also split across its historical entity, a post-2014 Addiko rebrand, and its Italian and Liechtenstein subsidiaries.
The two Austrian items were merged to restore full connectivity.
Within these constraints, the corrected graph captures major lifecycle milestones, a day-level coverage sequence around the regional fracture, personnel co-mention trajectories, and a difference in selected-relation density.

%% file: section_poland.tex
\section{Elite Networks in Polish News}
\label{sec:poland}\label{sec:application}

The Polish study analyses half a million news articles published between 1997 and 2025.
Before analysis, we merged name variants into Wikidata-verified canonical nodes and reviewed source articles to correct identity and relation-direction errors.
The resulting graph is heavy-tailed, with governing parties and state-linked institutions among its highest-degree entities.
We compare three structures with the historical record: the signed PO--PiS duopoly, state-owned-enterprise (SOE) patronage through overlapping governance and economic ties, and changes in polarisation across governing periods.
Since 2005, Polish party competition has operated as a duopoly between Civic Platform (PO, liberal-conservative) and Law and Justice (PiS, national-conservative), with Donald Tusk and Jarosław Kaczyński as the principal antagonists \autocite{markowski2006,tworzecki2019}.
Poland's major state-owned enterprises include Orlen, the state banks Powszechna Kasa Oszczędności Bank Polski (PKO~BP) and Pekao, and the energy firms Polska Grupa Energetyczna (PGE) and Polskie Górnictwo Naftowe i Gazownictwo (PGNiG).
They also include the insurer Powszechny Zakład Ubezpieczeń (PZU) and the miner KGHM Polska Miedź (KGHM).
These firms are documented sites of patronage where management turns over after changes of government, while business-political entanglement is more broadly documented in Poland \autocite{szarzec2022,mcmenamin2004}.
We analyse the central bank, Narodowy Bank Polski (NBP), separately as a state institution.
Following Olson's account of distributional coalitions, we treat governance--economic overlap around state-linked firms as an \emph{O-group} (rent-seeking) indicator.
The case study covers O-groups.
A Putnamian \emph{P-group} (civic-network) comparison requires separate indicators \autocite{olson1982,putnam1993,vonjacobi2025}.

We compute the network measures in Neo4j GDS over the deduplicated entity graph, excluding provenance edges and counting mentions by distinct article.
We aggregate ties using distinct-article counts and calculate patronage overlap strictly across QID-linked endpoints.

\subsection{Production Statistics}
\label{sec:pl-production}

Of \num{500000} sampled articles, \num{499999} were processed (one article exceeded the generation limit) and \num{410358} passed admission.
After cleanup, the graph contains \num{662411}~entity nodes and \num{1368851}~content relations across 101 relation types, spanning 34~news outlets.
The relation-level QID fill rate is \qtyrange{45}{49}{\percent}, well above the \qty{15.8}{\percent} node-level rate, because the high-degree entities that dominate the relation count are also the best linked.
Party nodes reach \qty{63.8}{\percent} coverage and politicians \qty{75.2}{\percent} once low-mention surface fragments (fewer than five mentions) are excluded.
Before macro-level aggregation, we merge name-keyed surface variants of twelve parties, the Warsaw Stock Exchange, the SOEs, and two national institutions into Wikidata-verified canonical nodes.
We also merge exact-name fragments of persons and organisations into the unique QID node of the same entity group, while other unlinked variants remain distinct.

We verified canonical entity identities against Wikidata and corrected or excluded mislinked nodes.

These 101 relation types belong to eight ontology groups.
For the signed analysis, we restrict the graph to QID-linked persons and parties joined by Governance or Stance relations.
Leiden remains unsigned, while the within- and cross-camp comparisons assign signs from relation type.

\subsection{Signed PO--PiS Cleavage}
\label{sec:pl-cleavage}

\begin{figure}[htbp]
	\centering
	\includegraphics[width=0.75\textwidth,height=65mm,keepaspectratio]{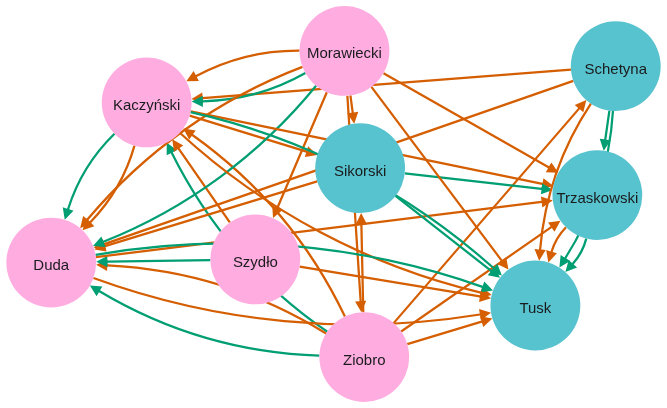}
	\caption{Extracted relations among nine front-line leaders in the PO--PiS cleavage.
		Cyan nodes identify the PO camp, and magenta nodes identify the PiS camp.
		Orange ties group negative relations (criticism, opposition and disputed statements), while green ties group positive relations (support).
		Each connection combines relations in both directions, so its arrowhead does not show who acted on whom.
		A sign is shown when at least two stored relations have that sign.}
	\label{fig:pl-cleavage}
\end{figure}

On the political subgraph (\num{5668} QID-linked persons and parties on Governance $\cup$ Stance edges), unsigned Leiden keeps the opposing leaders together at every $\gamma \in \{0.25, 0.5, 1, 2, 4\}$.
At $\gamma=1$, Leiden modularity \autocite{traag2019leiden} is 0.447, compared with 0.444 for Louvain \autocite{blondel2008louvain}, and the party \emph{organisations} split into families.
The front-line \emph{leaders} (Tusk, Kaczyński, Duda, Morawiecki, Ziobro, Trzaskowski, Hołownia, Kosiniak-Kamysz) remain in a single \enquote{conflict-core} community because unsigned modularity treats a \texttt{criticizes} edge like a \texttt{supports} edge.
We assign signs from the extracted relation type: \texttt{criticizes} and \texttt{opposes} are negative, while \texttt{supports} is positive (\Cref{fig:pl-cleavage}).
Within this subgraph, the signed comparison uses 18 curated persons assigned to five political camps, with party nodes excluded.
Between the PO and PiS camps, relations are \qtyrange{97}{99}{\percent} negative and symmetric (PiS$\to$PO 248 negative / 3 positive $\approx$ PO$\to$PiS 259 / 9).
Across all five camps, within-camp ties are 5.8$\times$ more likely to be positive than cross-camp ties (\qty{50.0}{\percent} vs.\ \qty{8.6}{\percent} positive), a pattern consistent with structural balance \autocite{cartwright1956balance}.
The internal camp structure remains multiplex, as allies accumulate both supportive and rivalrous mentions (the PO pair Tusk--Trzaskowski generates 20 positive against 2 negative edges).
The cleavage is defined by the resulting cross-camp asymmetry, with adversaries carrying far heavier edges than allies.
The Kaczyński--Tusk dyad is incident to 137 negative relations and 2 positive ones.

\begin{figure}[htbp]
	\centering
	\includegraphics[width=\textwidth]{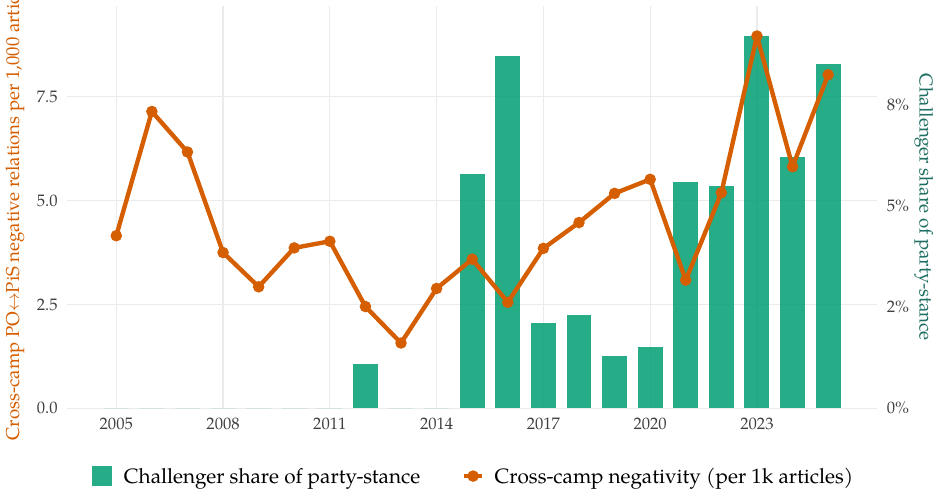}
	\caption{Article-normalised cross-camp antagonism and duopoly share over time.
		Cross-camp PO--PiS negativity per thousand articles rises 2.2$\times$ from its 2007--14 level to the 2023--25 peak, while the challenger share of party-level Stance activity rises from \qty{0}{\percent} to \qty{8.0}{\percent}.}
	\label{fig:pl-polarization}
\end{figure}

The unsigned Leiden partition also identifies four party families: PiS (838~nodes), PO and its Civic Coalition (770), the Left (654), and the Third Way / agrarian grouping (327).
It further identifies the conflict-core community of front-line leaders (423), a small Konfederacja community (108), and a separate community of foreign leaders (884).
These communities connect through the centrist pivots Szymon Hołownia and Władysław Kosiniak-Kamysz, the party-switchers Jarosław Gowin and Janusz Palikot, and foreign leaders (Putin, Trump, Merkel, Orbán) acting as shared reference points.
The network tracks this cleavage dynamically (\Cref{fig:pl-polarization}).
After normalising by article volume, cross-camp antagonism in the community-assigned camps rises 2.2$\times$ from 2007--14 to the 2023--25 KO era (3.5 to 7.6 negative PO--PiS relations per thousand articles), and 1.3$\times$ relative to 2005--06.
Relative to 2005--06, the negative-edge rate between persons assigned to the PO and PiS camps rises 4.6$\times$ (0.76 to 3.48 per thousand articles, or \qty{0.8}{\percent} to \qty{3.7}{\percent} of all negative edges between linked entities).
The duopoly's combined share of party-level Stance activity, which had grown from \qty{62}{\percent} to \qty{73}{\percent} during the 2015--22 PiS majority, falls to \qty{56.8}{\percent} in 2023--25.
Over the same period, the challenger share (Konfederacja, Polska~2050, Kukiz'15) rises from \qty{0}{\percent} to \qty{8.0}{\percent}, with the change beginning between 2022 and 2023.
The graph therefore combines higher cross-camp negativity with a larger challenger share.

\subsection{Network Topology}
\label{sec:pl-topology}

\begin{figure}[htbp]
	\centering
	\includegraphics[width=\textwidth]{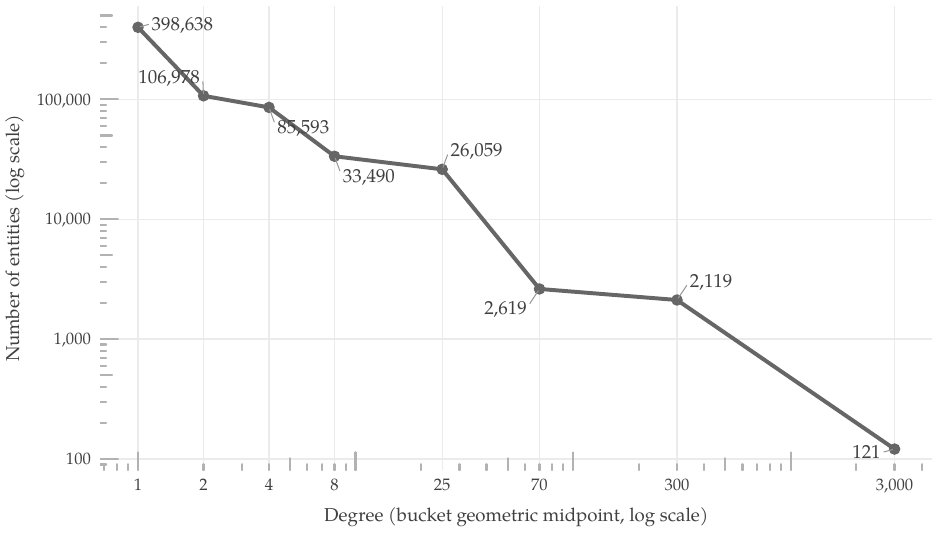}
	\caption{Degree distribution of the Polish entity graph (log-binned).
		\qty{60.2}{\percent} of entities have degree~1 and 121 exceed degree~\num{1000}.}
	\label{fig:pl-degree}
\end{figure}

\begin{figure}[htbp]
	\centering
	\includegraphics[width=\textwidth]{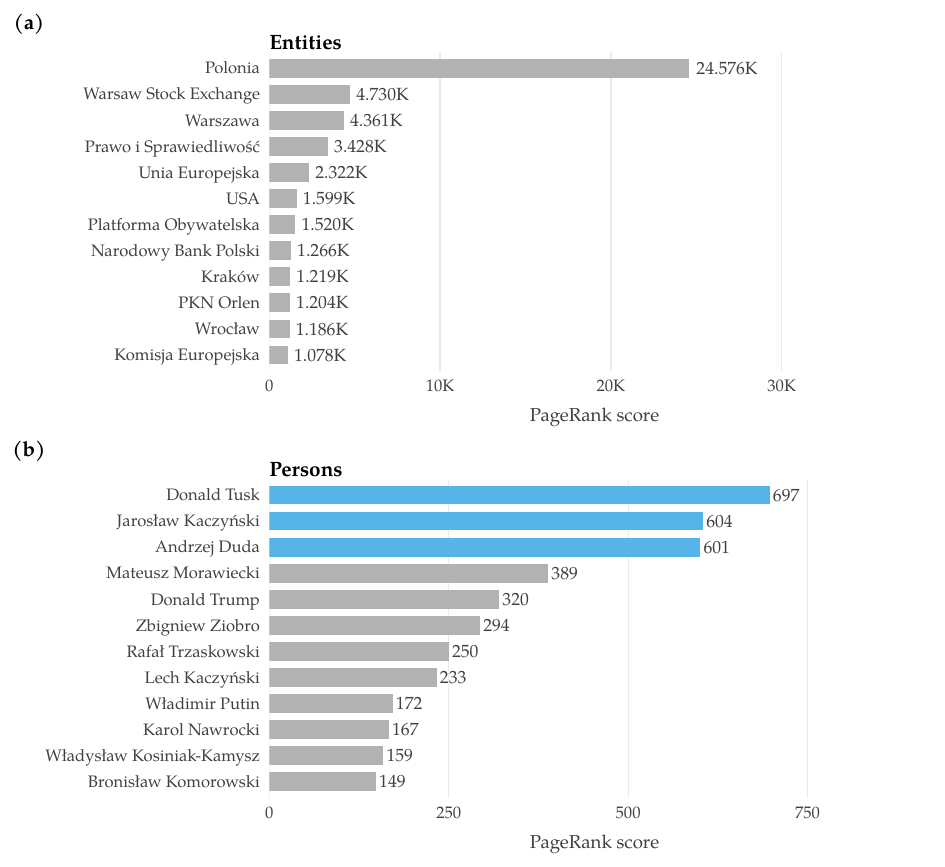}
	\caption{Weighted PageRank of the top entities \textbf{(a)} and top persons \textbf{(b)}.
		Tusk ranks most central among persons, with Kaczyński and Duda nearly tied behind him, while the NBP rises to \#8 among all entities.}
	\label{fig:pl-pagerank}
\end{figure}

Beyond the signed political subgraph, the full entity graph exhibits a sharp heavy-tailed degree distribution (\Cref{fig:pl-degree}): \num{398638} of \num{662411} entities (\qty{60.2}{\percent}) possess degree~1, while 121 exceed degree~\num{1000}.
The leading entities are Poland (Q36, degree~\num{45760}), Warsaw (\num{14218}), PiS (\num{13752}), the Warsaw Stock Exchange (GPW, \num{11294}), and the European Union (EU) (\num{8233}).
Every top-25 entity carries a valid QID and exceeds \num{2937}~mentions.
Among Polish actors, Donald Tusk leads the raw degree ranking (\num{3796}), followed by Andrzej Duda (\num{3237}) and Jarosław Kaczyński (\num{2522}), then Mateusz Morawiecki, Rafał Trzaskowski, and Zbigniew Ziobro.
Centrality leadership remains metric-dependent: Tusk leads under weighted PageRank \autocite{page1999pagerank} as well, but Kaczyński and Duda are within one percent of each other in second and third place, because the presidency structurally connects Duda to high-value state institutions.
The NBP ranks eighth among all entities under this measure (\Cref{fig:pl-pagerank}).
State-linked institutions dominate the high-degree organisation tier: Orlen (\num{6428}), PKO~BP (\num{3900}), PGNiG (\num{3727}), Pekao (\num{2986}), PZU (\num{2739}), PGE (\num{2551}), KGHM (\num{2482}), and the NBP (\num{1931}).
The ranking is consistent with the state-linked institutions identified in prior work.

\subsection{Relationship Patterns}
\label{sec:pl-relations}

\begin{figure}[htbp]
	\centering
	\includegraphics[width=\textwidth]{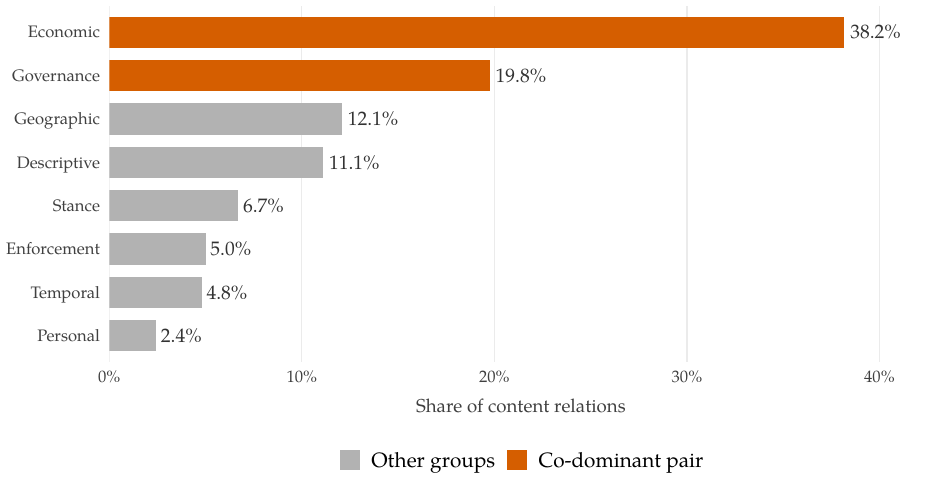}
	\caption{Distribution of relations across the eight ontology groups.
		Economic (\qty{38.2}{\percent}) and Governance (\qty{19.8}{\percent}) lead.}
	\label{fig:pl-groups}
\end{figure}

Across the eight extracted ontology groups, Economic (\qty{38.2}{\percent}, \num{522431}~edges) and Governance (\qty{19.8}{\percent}, \num{270409}) relations lead, trailed by Geographic (\qty{12.1}{\percent}) (\Cref{fig:pl-groups}).
The same ordering appears in both source classes: \qty{48.5}{\percent} Economic against \qty{20.9}{\percent} Governance in the financial wires, and \qty{34.5}{\percent} against \qty{19.7}{\percent} in political outlets.
The \texttt{employer} relation is the single most frequent type (\qty{11.2}{\percent}), followed by \texttt{position\_held} (\qty{6.2}{\percent}).

\begin{figure}[htbp]
	\centering
	\includegraphics[width=\textwidth]{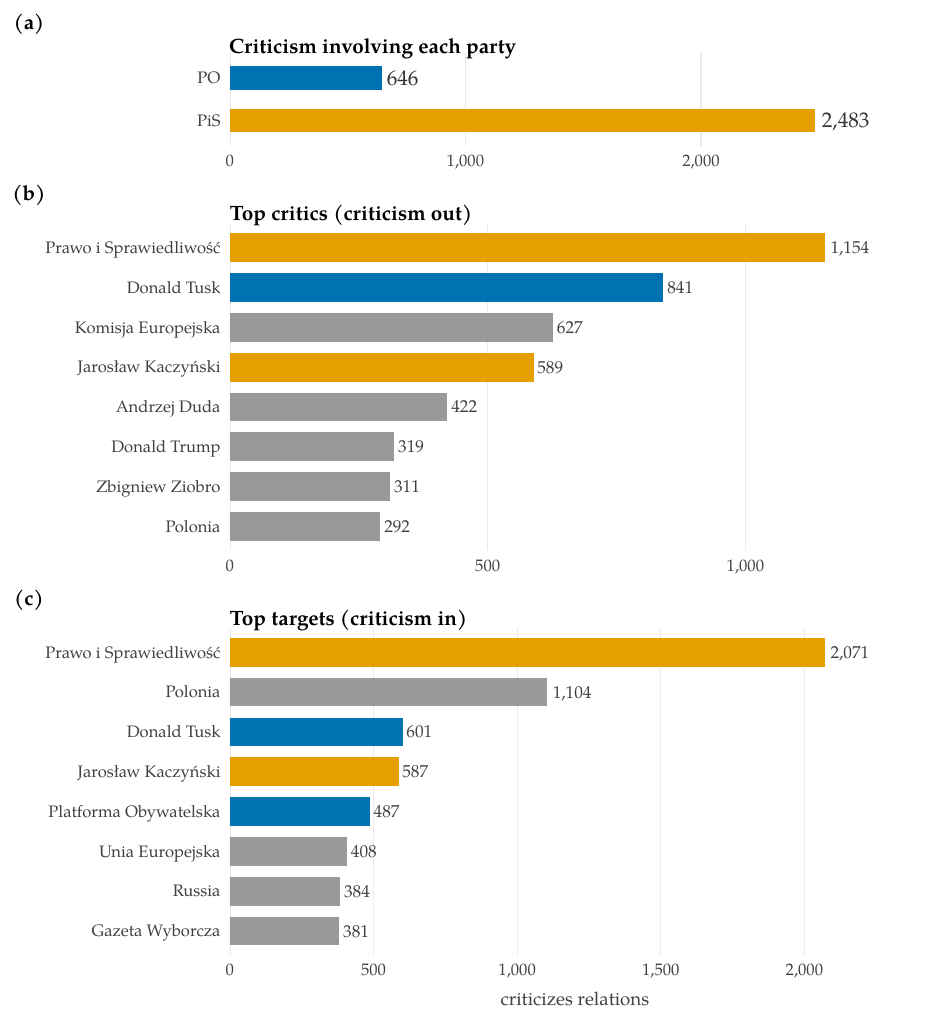}
	\caption{Extracted \texttt{criticizes} relations in Polish news.
		\textbf{(a)}~Relations involving each party and another QID-linked entity, counting both incoming and outgoing criticism.
		\textbf{(b)}~Top critics (largest out-degree).
		\textbf{(c)}~Top targets (largest in-degree).
		Bars are coloured by formal party membership: orange for PiS, blue for PO, grey otherwise.
		Andrzej Duda (independent) and Zbigniew Ziobro (Suwerenna Polska) appear grey as non-members of PiS.}
	\label{fig:pl-criticizes}
\end{figure}

Extracted partisan conflict is sharply directional (\Cref{fig:pl-criticizes}).
PiS is incident to \num{2483} \texttt{criticizes} edges with a QID-linked counterpart against PO's 646, yielding a $3.84{:}1$ ratio (bootstrap \qty{95}{\percent} CI $[3.53, 4.20]$).
PiS also absorbs more criticism than it emits (in-degree \num{2071} vs.\ out-degree \num{1154} over all counterparts) and is the corpus's largest single criticism target.
The observed period includes its eight-year government tenure (2015--23) and the 2024 government transition.
Among persons, Tusk (841 edges) and Kaczyński (589) are the most-criticising actors and also the most-criticised (601 and 587 incoming edges), while PiS as a party absorbs \num{2071} edges.
Party-label-to-party-label criticism is sparse (372 edges, 59 in each direction between PiS and PO): conflict operates at the person-to-person and person-to-party levels.
Annual relation volume sustains a high plateau throughout the 2015--2023 PiS majority before peaking during the 2024 government transition (\num{99474}~relations, \num{28214}~articles).
Because per-article relation intensity stays within 2.7--3.9 across the timeline, the 2024 surge is an article-volume peak.
The temporal antagonism rates below therefore normalise by total article volume.

\subsection{Firm-Level O-Group}
\label{sec:pl-multiplex}

\begin{figure}[htbp]
	\centering
	\includegraphics[width=\textwidth]{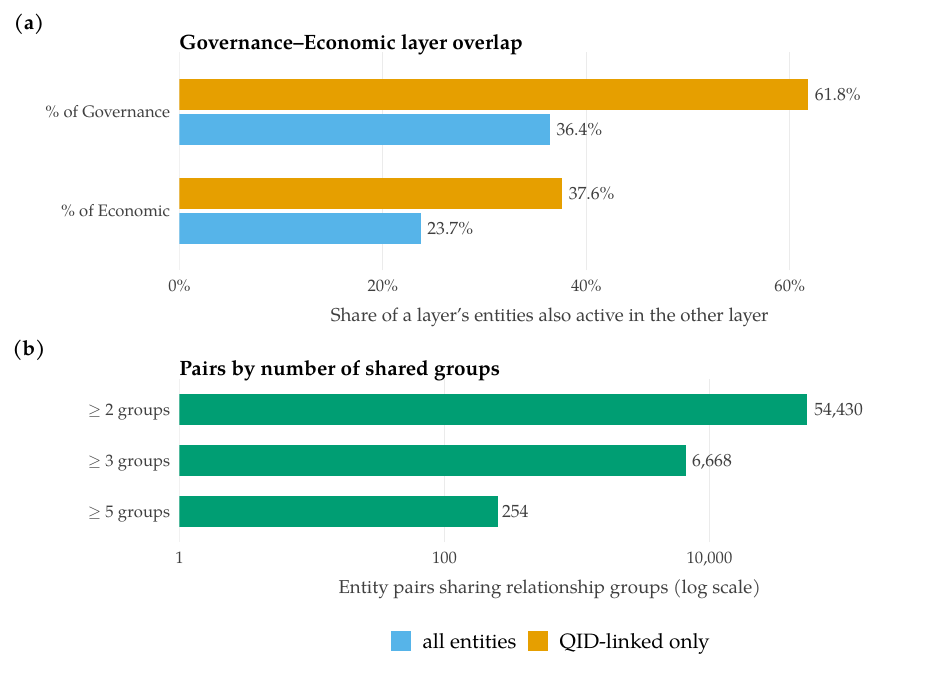}
	\caption{Governance$\cap$Economic multiplex overlap among entities and relation pairs.
		\textbf{(a)}~Share of each layer's entities also active in the other layer (\qty{36.4}{\percent} of governance and \qty{23.7}{\percent} of economic participants).
		\textbf{(b)}~Multiplex depth: entity pairs sharing $\geq 2$, $\geq 3$, and $\geq 5$ relationship groups.}

	\label{fig:pl-multiplex}
\end{figure}

The graph shows governance$\cap$economic overlap (\Cref{fig:pl-multiplex}).
Of the entities participating in Governance and Economic relations, \num{74406} appear in both (\qty{36.4}{\percent} of governance and \qty{23.7}{\percent} of economic participants).
Restricted to QID-linked nodes, the overlap rises to \qty{61.8}{\percent} and \qty{37.6}{\percent}, although this restriction also selects better-linked entities.
Many entity pairs connect through several relationship groups: \num{54430} through two or more, \num{6668} through three, and 254 through five or more.
Named entities generate \qty{62}{\percent} of the $\geq$3-group pairs and \qty{90}{\percent} of the $\geq$5-group pairs.
Among \num{1173} distinct economic ties between political or official actors and firms or related institutions, state-owned or state-linked entities have the highest domestic counts.
The NBP (83~distinct political or official actors), PKO~BP, Orlen, PGNiG, Bank Gospodarstwa Krajowego (BGK), and PGE lead the domestic distribution.
The state-linked core ranks above private firms.
The other high-count entities are foreign central banks and the European Central Bank, which reflect monetary-policy coverage.
Among entities with Economic degree of at least 200, the thirteen predefined SOEs have a higher median ratio of distinct political or official neighbours to Economic degree than the 120 non-SOEs (0.0053 vs 0.0025).
The seven-group pairs reflect state-to-state diplomacy and rating agency activity, including Poland$\leftrightarrow$European Commission (983~edges), Poland$\leftrightarrow$Russia (188), and Poland$\leftrightarrow$Standard~\&~Poor's (170).
The sole domestic exception is the Orlen$\leftrightarrow$Lotos merger pair (550~edges).
Polish patronage is nonetheless often multiplex: Orlen's 2020--2021 acquisition of the Polska~Press regional-media group connects the economic and media-governance layers.
The firm-level pattern is therefore consistent with the O-group expectation, while most of the extreme multiplex tail reflects international diplomacy.

\subsection{SOE Patronage Turnover}
\label{sec:pl-turnover}

\begin{figure}[htbp]
	\centering
	\includegraphics[width=\textwidth]{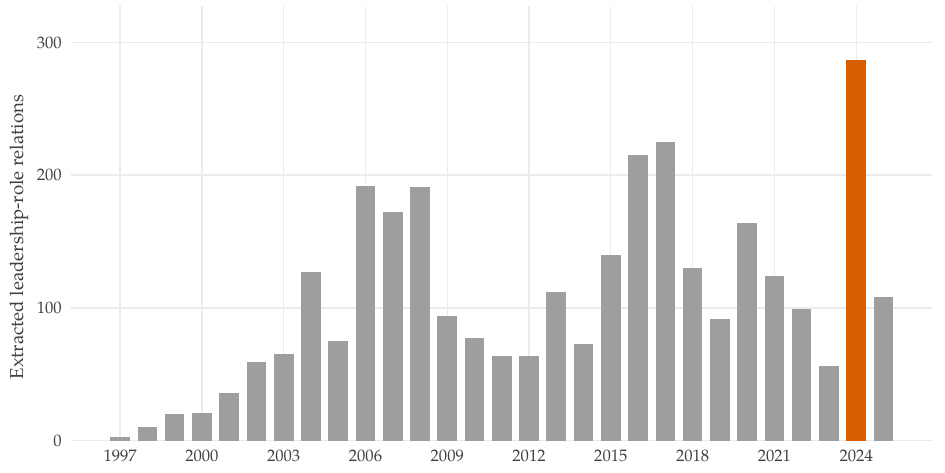}
	\caption{Annual extracted leadership-role relations at major Polish SOEs (raw edge count).
		The highlighted 2024 bar (287 edges, the largest in the series) follows the 12.2023 inauguration of the KO-led coalition government.}
	\label{fig:pl-turnover}
\end{figure}

The extracted ties between state-linked institutions and political actors are typed role relations.
Daniel Obajtek's Orlen edges concentrate in his 2018--2024 CEO term, apart from three edges dated earlier.
Adam Glapiński's NBP edges begin with his 2010--2016 Monetary Policy Council membership and concentrate after he became governor in 2016.
The relation types record role occupancy, including \path{chief_executive_officer}, \path{supervisory_board_member}, and \path{dismissed_from}, and the most prominent ties carry high source counts (Glapiński's NBP tie, $n=498$).
The network routes these ties through individuals and through bodies that exercise state ownership, including the State Treasury, the Ministry of Finance, and the development fund Polski Fundusz Rozwoju (PFR).
Direct SOE$\leftrightarrow$party edges are rare (78 in the whole graph, 30 of them to PiS).
Extracted role-relation counts provide a marker of leadership turnover (\Cref{fig:pl-turnover}).
The largest annual count occurs in 2024 (287 role-relation edges, compared with 225 in 2017), following the December 2023 inauguration of the Civic Coalition (Koalicja Obywatelska, KO)-led government.
The 2024 total includes 123 dismissal edges across 70 persons, led by Orlen (41), KGHM (17), Tauron (15), PGE (14), and PZU (13), and encompasses Obajtek (Orlen) and the PGE and Jastrzębska Spółka Węglowa (JSW) boards.
Because these counts peak around government changes, we report the 2024 increase as a single event and fit no trend to the series.
The NBP differs from government-replaceable firms: it attracts the most distinct political persons of any state-linked institution (199), while its constitutional status and protected six-year gubernatorial term require separate interpretation \autocite{polishconstitution1997}.
The SOE-patronage claim therefore rests on the government-replaceable firms: Orlen, PKO~BP, PGE, PZU, KGHM, and BGK.

\subsection{Validation and Limits}
\label{sec:pl-validation}

The corrected graph contains the expected PO--PiS cleavage, state-linked institutional prominence, and governance--economic overlap.
This correspondence provides face validity for the selected structures and is consistent with the O-group expectation.

The analysed graph includes targeted merging of name variants, source review, and direction and identity corrections.
Extracted party-membership relations sometimes assign opponents to PiS when parsing formulations such as \enquote{X criticised Y's party}.
The signed analysis therefore uses the fixed 18-person camp map and excludes extracted membership edges.
Community-assigned analyses retain those assignments and classify remaining nodes by the $\gamma=1$ Leiden community containing each party anchor.
The QID-restricted analyses also select better-linked entities, and the turnover totals count extracted role-relation edges, which can exceed the number of unique leadership changes.